\documentclass[journal]{IEEEtran}
\usepackage[usenames]{color}
\usepackage{epsfig}
\usepackage{graphics}
\usepackage{amsmath}
\usepackage{amssymb}
\usepackage{multirow}
\usepackage{cite}
\usepackage{array}
\usepackage{pslatex} 
\usepackage{url}
\usepackage{caption}
\usepackage{lineno}
\usepackage{graphicx}  % Written by David Carlisle and Sebastian Rahtz
\usepackage{setspace}
\usepackage{tikz}
\usepackage{hhline}
\usepackage{mathtools}
\usepackage[letterpaper]{geometry}
\ifCLASSINFOpdf
\else
\fi
\usepackage[english]{babel}
\usepackage[utf8]{inputenc}
\usepackage{fancyhdr}
\begin{document}
\title{Adaptation of Autoencoder for Sparsity Reduction From Clinical Notes Representation Learning\vspace{0.25cm}}

%\textcolor{{violet}


\author{Thanh-Dung Le,~\IEEEmembership{Member,~IEEE,}
        Rita Noumeir Ph.D.,~\IEEEmembership{Member,~IEEE,}\\
      Jérôme Rambaud M.D., Ph.D., 
      Guillaume Sans M.D., and Philippe Jouvet M.D., Ph.D.
      % <-this % stops a space
\thanks{This work was supported in part by the Natural Sciences and Engineering Research Council (NSERC), in part by the Institut de Valorisation des données de l’Université de Montréal (IVADO), in part by the Fonds de la recherche en sante du Quebec (FRQS), and in part by the Fonds de recherche du Québec – Nature et technologies (FRQNT). 
	\par Thanh-Dung Le is with the Biomedical Information Processing Lab, \'{E}cole de Technologie Sup\'{e}rieure, University of Qu\'{e}bec,  Montr\'{e}al, Qu\'{e}bec, Canada, and also with the CHU Sainte-Justine Research Center, CHU Sainte-Justine Hospital, University of Montreal, Montr\'{e}al, Qu\'{e}bec, Canada (Email: thanh-dung.le.1@ens.etsmtl.ca).
	\par  Rita Noumeir is with the Biomedical Information Processing Lab, \'{E}cole de Technologie Sup\'{e}rieure, University of Qu\'{e}bec,  Montr\'{e}al, Qu\'{e}bec, Canada. 
	\par Jérôme Rambaud, Guillaume Sans, and Philippe Jouvet are with the CHU Sainte-Justine Research Center, CHU Sainte-Justine Hospital, University of Montreal, Montr\'{e}al, Qu\'{e}bec, Canada.
}}

\maketitle\thispagestyle{fancy}

\begin{abstract}
When dealing with clinical text classification on a small dataset recent studies have confirmed that a well-tuned multilayer perceptron outperforms other generative classifiers, including deep learning ones. To increase the performance of the neural network classifier, feature selection for the learning representation can effectively be used. However, most feature selection methods only estimate the degree of linear dependency between variables and select the best features based on univariate statistical tests. Furthermore, the sparsity of the feature space involved in the learning representation is ignored.  \textit{Goal}: Our aim is therefore to access an alternative approach to tackle the sparsity by compressing the clinical representation feature space, where limited French clinical notes can also be dealt with effectively. \textit{Methods}: This study proposed an autoencoder learning algorithm to take advantage of sparsity reduction in clinical note representation. The motivation was to determine how to compress sparse, high-dimensional data by reducing the dimension of the clinical note representation feature space. The classification performance of the classifiers was then evaluated in the trained and compressed feature space.  \text{Results}: The proposed approach provided overall performance gains of up to 3\% for each evaluation. Finally, the classifier achieved a 92\% accuracy, 91\% recall, 91\% precision, and 91\% f1-score in detecting the patient’s condition. Furthermore, the compression working mechanism and the autoencoder prediction process were demonstrated by applying the theoretic information bottleneck framework.
%Strengths and weaknesses of the classifier were also presented in cases where the number of training examples was increased.
\end{abstract}


% \fi
\IEEEpeerreviewmaketitle

%\begin{minipage}[t]{1\columnwidth}
\textbf{\textit{Impact Statement-} Autoencoder effectively tackles the problem of sparsity in the representation feature space from a small clinical narrative dataset.  Significantly, it can learn the best representation of the training data because of its lossless compression capacity as compared to other approaches. Consequently, its downstream classification ability can also be significantly improved, which cannot be done using deep learning models.}\\
\\
%\end{minipage}

\section{INTRODUCTION}

\IEEEPARstart{C}{linical} decision support systems (CDSS) are continuously being developed and play a crucial role in promoting a personalized healthcare system, as more and more data are collected and stored continuously \cite{musen2021clinical}. These data represent decisive points in advancing and enhancing the efficiency and effectiveness of CDSS operations. Predictive models have been developed based on the latter for preventive treatment and patient diagnosis, culminating in intelligent, precise, and timely healthcare improvement \cite{sutton2020overview}. In one notable example, a recent study \cite{gold2022effect} analyzed the effect of CDSS on cardiovascular risk in 18,578 patients in 70 community health centers. In that case, CDSS significantly reduced the risk of cardiovascular disease among vulnerable high-risk patients. 

\begin{figure*}[!t]
	\centering
	\vspace{2pt}
	\includegraphics[scale=0.680]{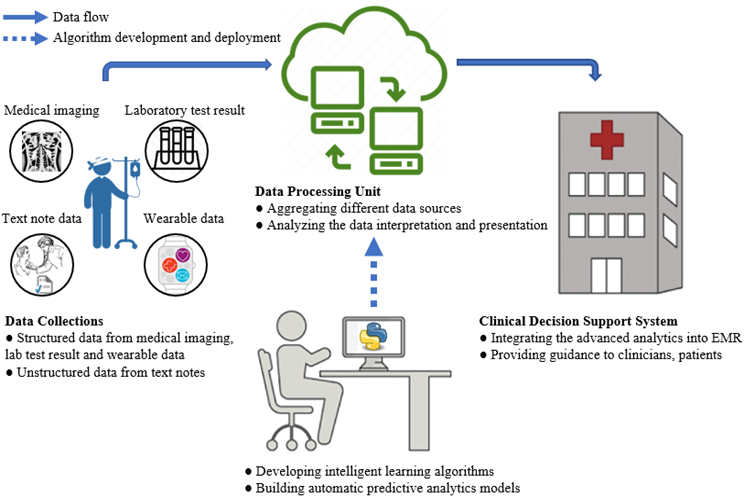}
	\vspace{-2pt}
	\caption{Workflow demonstration of a clinical decision-support system at CHU Sainte-Justine hospital.}
	\vspace{-5pt}
	\label{fig:CHUSJ_workflow}
\end{figure*}

Following the above successes, a CDSS was developed at CHU Sainte-Justine Research Center (CHUSJ). The system monitors the management of pediatric intensive care for all patients ranging in age from 0 to 18. Fig. \ref{fig:CHUSJ_workflow} illustrates two fundamental processes in the CDSS workflow at CHUSJ, which involve collecting and processing critical care data. First, clinical data are collected and stored in a clinical data warehouse. The data processing unit is then systematically aggregated and processed to convert raw data to a machine-readable form in the data processing unit. This process helps analyzing the unknown data interpretation and presentation. The CDSS can thus integrate the advanced analytic result of the data processing unit and learning algorithms; then clinicians can adequately use the CDSS to guide early intervention and prevention for healthcare management. 

One of the goals of the CDSS system in CHUSJ is automatically screening the data from electronic medical records, chest X-rays and other data sources, which has the potential to increase the diagnosis rate and then improve the management of acute respiratory distress syndromes (ARDS) in real time. Usually, the diagnosis of ARDS was delayed or missed in two-thirds of patients, and the diagnosis was missed completely in 40\% of patients \cite{bellani2016epidemiology}. To make the diagnosis of ARDS, three main conditions need to be detected: hypoxemia (low blood oxygenation), presence of infiltrates on chest X Ray and absence of cardiac failure \cite{pediatric2015pediatric}. Our research team has developed algorithms for hypoxemia \cite{sauthier2021estimated}, chest X-ray analysis \cite{zaglam2014computer}, and identification of the absence of cardiac failure \cite{le2021detecting, le2021machine1}. Technically, we successfully performed extensive analyzes of machine learning algorithms (ML) aimed at detecting cardiac failure from clinical narratives using natural language processing based on such algorithms \cite{le2021detecting}. The study included the clinical notes of 1386 patients classified by two independent physicians using a standardized approach. It confirmed that the framework proposed  herein yields an overall classification performance with 89\% accuracy, 88\% recall, and 89\% precision by applying a multilayer perceptron neural network (MLP-NN) classifier in combination with a term frequency × inverse document frequency (TF-IDF) learning representation for clinical notes.

These results were made possible by the contributions of the feature selection process, also known as SelectKBest. The advantage of the process was proven for supervised models as the classifier performance brought overall improvements of up to 3-4\% over the case without the feature selection. It is obvious to understand because there are fewer misleading features; after selecting the best K features, the classifier accuracy effectively improves. Unfortunately, the SelectKBest feature selection continues to have certain limitations in the proposed framework. One reason is that the feature selection method is based on a statistical test that estimates the degree of linear dependency between random variables. Then, it removes irrelevant features and ignores the correlation between data elements. As a result, more samples are required for an accurate estimation and avoidance of overfitting, which is not possible in our case \cite{jain2018feature}. Furthermore, SelectKBest does not deal mainly with the sparsity of the feature space in the note representation matrix \cite{forman2003extensive}. Consequently, the sparsity that characterizes the learning representation vector space is ignored.  

 %With the high sparsity ratio obtained from the TF-IDF representation feature space, the SelectKBest effectively increases the classifier's performance \cite{le2021detecting}. 

In the health care field, the autoencoder algorithm (AE) has lived up to its promises and has shown its effectiveness when it comes to improving outcomes for efficient clinical decision making. First, AE can find informative transformed feature vectors through the compressed latent representation. For example, a study \cite{zhou2018optimizing} demonstrates an efficient framework for automatically learning compact representations from heterogeneous raw data sources from patient health data. In addition, AE can improve the predictability of the six different learning models to detect Parkinson's classification \cite{xiong2020deep}. Another study \cite{kolyvakis2018biomedical} shows that AE improved the performance of a novel outlier detection mechanism by retrofitting word vectors for the biomedical ontology matching task. Second, having rich and accurate clinical data is very challenging \cite{quiroz2019challenges} because the acquisition and sharing of medical data faces a significant obstacle in the form of privacy issues and the sensitive nature of the data. Fortunately, AE can be applied for sparsity reduction in clinical  representation feature space to allow to tackle problems related to limited data availability. It could effectively discover the low dimensional embeddings, and reveal the underlying effective manifold structure from a sparse high dimensional document-term matrix \cite{leyli2017denoising}.

Therefore, the present study examines alternatives to feature selection and focuses mainly on compressing data without loss of information by employing an AE algorithm. First, we aim to achieve a better feature space without any sparsity. We are interested in compressing the sparse TF-IDF matrix and reducing its dimensions to improve the efficiency of the feature space representation. Notably, we incorporate a neural network to learn efficient codings of unlabeled data to address the issues caused by sparse vectors generated from the TF-IDF representation feature space for clinical notes. Then, the compressed vector space from the TF-IDF matrix is fed into the classifiers as a refined input. Finally, ML classifiers conduct the learning process to draw comparative results, which are then used to evaluate performance for the classification task.

%Second, The motivation is to adequately compress the sparse, high-dimensional data by reducing the dimension from the feature space of the TF-IDF representation matrix.

Our study confirms that AE effectively compresses the vector space of the TF-IDF representation for clinical narratives into a lower dimension. The proposed approach can retain the critical feature by capturing the correlation between attributes during the training process, hence; the downstream classification task can generally be increased to 2-3\% for each evaluation criterion. Furthermore, we also highlight the value of AE behaviors in a limited data set. We analyze the working mechanism of the AE, which explains how the AE works to compress data through the encoder and decoder. Based on the information-theoretic framework, the working mechanism of the AE is to optimize the information bottleneck during the compression and prediction process, respectively. As a result, the behavior of AE in limited data is exactly in harmony with such cases where there are much larger data availability. 

Section \ref{sec:autoencoder} will discuss the materials and methods. The experimental results and discussion then will be discussed in section \ref{sec:exp_result}, \ref{sec:dis}. Finally, section \ref{sec:conclusion} provides concluding remarks.

\section{MATERIALS AND METHODS}
\label{sec:autoencoder}

\subsection{Data Sparsity Challenges}

%From our recent study \cite{le2021detecting}, the combination of TF-IDF and MLP-NN consistently outperforms other combinations with overall performance and is the most stable in all circumstances. Without any feature selection, the proposed framework yielded an overall classification performance with acc, pre, rec, and f1 of 85\% and 84\%, 85\%, and 84\%, respectively. Also, if the feature selection (SelectKBest) was well applied and tuned, it could improve up to 3-4\% for each evaluation in the overall performance. Consequently, it achieves the best performance with 89\%, 89\%, 88\%, and 88\% for acc, pre, rec, and f1, respectively. 

In numerical analysis, a sparse matrix or sparse array is a matrix in which most elements are zero \cite{hurley2009comparing}. The number of zero-valued elements divided by the total number of elements (e.g., $m \times n$ for a $m \times n$ matrix) is called the matrix sparsity (equal to 1 minus the density of the matrix). Using these definitions, a matrix will be sparse when its sparsity is more significant than 0.5. In our case, we have more than 580000 (n-grams) word count from 5444 single lines of notes with 1941 positive cases (36\% of total) and 3503 negative cases. We applied the SelectKBest to select top best `k=20000' of the vectorized features for the TF-IDF preresentation learning feature space. Finally, we have a matrix of features of $(5444 \times 20000)$. With the sparsity, it is calculated by the Eq. \ref{eq: sparse_cal}, and the sparsity of the TF-IDF matrix is greater than 0.9.

It confirms that the representation matrix from the TF-IDF is sparse because every word is treated separately. Hence, the semantic relationship between separated entities is ignored, which would cause information loss. Although the combination of TF-IDF and MLP-NN consistently outperformed other combinations with overall performance and was the most stable under all circumstances \cite{le2021detecting}, the sparsity remains. Therefore, the motivation is how we can compress the sparse, high-dimensional data by reducing the dimension from the feature space of clinical notes representation TF-IDF feature space.

\begin{align}
    \text{sparsity}= 1 - \frac{\text{count\_nonzero(TF-IDF)}}{\text{total\_elements\_of\_(TF-IDF)}}
    \label{eq: sparse_cal}
\end{align}

\subsection{Autoencoder Learning Algorithm}
An AE was originated by \cite{kramer1991nonlinear} to solve a nonlinear dimensional reduction; later AE was famously promoted by training a MLP-NN with a small central layer to reconstruct high-dimensional input vectors \cite{hinton2006reducing, wang2016auto}. Technically, AE takes an input $X \in \mathcal{R}^{N \times D}$ and, maps it to a latent representation $Z \in \mathcal{R}^{N \times M}$ via a nonlinear mapping. Let us call $x \in X$, and $z \in Z$, we will have:

\begin{align}
    z=g(Wx+b)
\end{align}

\noindent $W$ is a weight matrix during training, $b$ is a bias vector, and $g(\cdot)$ stands for a nonlinear function, such as the logistic sigmoid function or a hyperbolic tangent function. The encoded feature representation $x$ is then used to reconstruct the input $x$ by reverse mapping, leading to the reconstructed input $x'$:

\begin{align}
 x' = f(W'z+b')
\end{align}

\noindent where $W'$ is usually limited to the form of $W'=W^T$, i.e. the same weight is used to encode the input and decode the latent representation. $f(\cdot)$ is also a non-linear function. The AE tries to learn a function $f_{W', b'}(x) \approx x'$. In other words, it is trying to learn an approximation of the identity function for the output $x'$ that is similar to $x$. Still, by placing constraints on the network, such as limiting the number of hidden units, we can discover interesting data structures. Then, the reconstruction error is defined as the Euclidean distance between $x$ and $x'$ that is constrained to approximate the input data $x$ (that is, minimizing $||x-x'||^2$). 
\begin{align}
    \mathcal{L}\left(x, x'\right) &=\left\|x-x'\right\|^2 \nonumber \\ 
                                  &=\left\|x-f(W'\left(g(Wx+b) \right)+b') \right\|^2
    \label{eq:reconstruction_error}
\end{align}

For the reconstruction evaluation between the original data x, and the reconstructed output $x'$, we will apply the statistical measure $R^2_i$ for the $i^{th}$ variable of $x_i$. It can be computed as: 

\begin{align}
    R^2_i = 1 - \frac{\sum_{j=1}^m (x_{j,i} - x'_{j,i})^2}{\sum_{j=1}^m x_{j,i}^2}
    \label{eq:reconst_eval}
\end{align}

\noindent Since $R^2=1$ will be a perfect reconstruction; we will evaluate the reconstruction by how much the value of $R^2$ is close to 1.

Ideally, an effective AE can be designed and trained based on the minimization of reconstruction error from Eq. \ref{eq:reconstruction_error} and maximization of the reconstructed effectiveness from Eq. \ref{eq:reconst_eval}; however, it is substantially based on its width (number of neuron units or latent representation dimension $M$) and its depth (number of hidden layers). First, conventional AE relies on the dimension of the latent representation $z$ being smaller than that of the input $x$ ($M<D$), which means that it tends to learn a low-dimensional compressed representation. The study \cite{garg2020functional} presents methods to learn the decoder function $f(\cdot)$ as a learnable function through the reconstruction error in Eq. \ref{eq:reconstruction_error} in several representation learning approaches. It is concluded that the compression depends on dimension $M$ but less on dimension $D$. Second, it has been shown that training a neural network-based by increasing the number of hidden layers (in combination with an increase in the number of neuron units per layer) achieves less consistent results \cite{steinmeyer2020sampling}. Therefore, in our case, we use a small and simple AE. We employ an AE with three layers (one input layer, one hidden layer, and one output layer). Mainly, to reduce the parameters from the latent space of the AE, we apply the regularization technique from study \cite{shi2019learning} to remove redundant parameters.  

%the AE is affected more on its latent representation $M$ than the data dimension $D$. C

\begin{figure}[t]
	\centering
	\vspace{2pt}
	\includegraphics[scale=0.5]{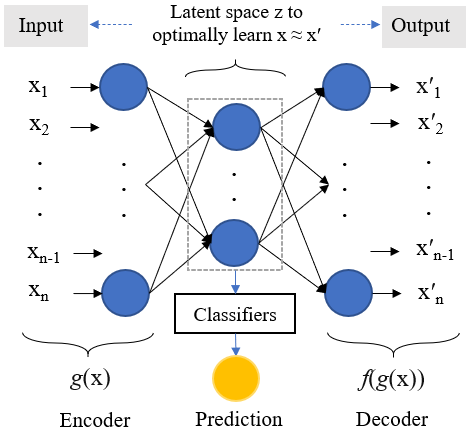}
	\vspace{-2pt}
	\caption{Schematic structure of an AE-based for compression and prediction.}
	\vspace{-5pt}
	\label{fig:autoencoder_schem}
\end{figure}

After training, we use the weight matrix from the hidden layer as a pre-trained tool. A classifier subsequently use this pre-train latent space representation to perform the binary classification, as shown in Fig. \ref{fig:autoencoder_schem}. For the classifiers, it is essential to have consistency in evaluating the proposed approach's performance. Then, we employ six different ML classifiers, including Random Forest (RF), Multinomial Naive Bayes (MultinomialNB), Logistic Regression (LR), Support Vector Machine (SVC), Gaussian Naive Bayes (GaussianNB) and Multilayer Perceptron Neural Network (MLP-NN).

Furthermore, to understand the dynamics of learning and the behavior of AE; especially, in our case with limited data, we also analyze the behavior of AE during the training process from the encoder and decoder. Technically, we capture to understand how the AE can retain the information during the compression process. To do that, we apply the information-theoretic quantities and their estimators based on information-theoretic learning, which compute and optimize information-theoretic descriptors named mutual information. The information-theoretic framework \cite{yu2019understanding, tapia2020information, lee2021information} has been utilized for a detailed theoretical explanation of an AE. These studies rely on the ``information bottleneck'' \cite{tishby2000information, shwartz2017opening} to understand and estimate how the AE works by quantifying its information plane coordinates. The information bottleneck can be used as an optimal bound that maximally compresses the input $x$, for a given mutual information on the desired output $x'$. There are comprehensive overviews of recent studies \cite{geiger2021information, geiger2020information, alomrani2021critical}. Technically, we first bin the output activation as stated in \cite{shwartz2017opening}, and we treat each hidden layer $i$ ($1 \leq i \leq K $) as a single variable $T_i$. Then we will be able to estimate the mutual information between all the hidden layers and the input/output layers by estimating the joint distribution $P(X,T_i)$ and $P(T_i,X')$, and use them to calculate the mutual information of the encoder (between the input $X$ and the hidden layer $T_i$), and the mutual information of the decoder (between the hidden layer $T_i$ and the desired output $X'$) using the following equations Eq. \ref{eq:ib1}, \ref{eq:ib2}. Finally, we can learn the good representation $T(X)$, which is characterized by its encoder and decoder distribution $P(T|X)$, and $P(X'|T)$, respectively, to effectively map the input patterns $X$ to a good prediction of the desired output $X'$.

\begin{align}
    I(X;T_i) = \sum_{x\in X, t\in T_i}P(x,t)\log\Big(\frac{P(x,t)}{P(x)P(t)}\Big)
    \label{eq:ib1}
\end{align}
    
\begin{align}
    I(T_i;X') = \sum_{t\in T_i, x'\in X'}P(t,x')\log\Big(\frac{P(t,x')}{P(t)P(x')}\Big).
    \label{eq:ib2}
\end{align}

%\cite{krishnan2018challenges, cemgil2020autoencoding}

\section{Results}
\label{sec:exp_result}

To assess the performance of our method, metrics including accuracy, precision, recall (or sensitivity), and F1 score were used \cite{goutte2005probabilistic}. These metrics are defined as follows. 
\begin{align}
&\text {Accuracy (acc) }=\frac{\mathrm{TP}+\mathrm{TN}}{\mathrm{TP}+\mathrm{TN}+\mathrm{FP}+\mathrm{FN}} \nonumber \\ 
&\text {Precision (pre) }=\frac{\mathrm{TP}}{\mathrm{TP}+\mathrm{FP}} \nonumber \\
&\text {Recall/Sensitivity (rec)}=\frac{\mathrm{TP}}{\mathrm{TP}+\mathrm{FN}} \nonumber \\ 
&\text {F1-Score (f1)} =\frac{2^{\star} \text {Precision}^{\star} \text {Recall}}{\text {Precision }+\text {Recall}} \nonumber
\end{align}

\noindent where TN and TP stand for true negative and true positive, respectively, and are the number of negative and positive patients correctly classified. FP and FN represent false positives and false negatives, respectively, and represent the number of positive and negative patients incorrectly predicted.

For implementation, we used the same hyperparameters from our previous study \cite{le2021detecting} for all classifiers so that we can have a consistent evaluation for the performance. The data was also divided into 60\% training, 20\% validation, and 20\% testing. The implementation was done using Python Scikit learn \cite{pedregosa2011scikit} and Keras \cite{chollet2015keras}.  We performed a grid search for up to three hidden layers and 500 neurons per layer, and other hyperparameters are summarized in Table \ref{tab:my-ae_hyp} for AE training. For the optimizers, we used the Stochastic Gradient Descent (SGD) and  Adaptive Moment Estimation (ADAM) with small scalar $\epsilon$, and the forgetting factors for gradients and second moments of gradients, $\beta_1$ and $\beta_2$, respectively. Then, a combination with the highest estimations was considered the best performance. 

\begin{table}[!t]
\centering
\footnotesize
\vspace{2pt}
\caption{ Hyperparameters Summary for AE Trainning}
\label{tab:my-ae_hyp}
%\resizebox{\textwidth}{!}{%
\begin{tabular}{|l|l|}
\hline
\textbf{Hyperparameter}     & \textbf{Ranges}        \\ \hline
Hidden layers      & 1-3           \\ 
Neurons            & 100-500        \\ 
Activation         & Sigmoid \\ 
Kernel initializer & GlorotNormal  \\ 
Optimizers         & SGD,  ADAM      \\ 
Learning rate      & 0.001 - 0.01  \\ 
$\beta_1$            & 0.9           \\ 
$\beta_2$            & 0.999         \\ 
$\epsilon $           & $e^{-8}$ - $e^{-7}$   \\ \hline
\end{tabular}%
\vspace{-5pt}
%}
\end{table}

\section{Discussion}
\label{sec:dis}
To deal with the sparsity, many researchers simply focus on dimension reduction. There are two most popular techniques, namely Linear Discriminant Analysis (LDA) and Principal Component Analysis (PCA), for their simplicity among other dimension reduction techniques \cite{anowar2021conceptual}, even with a large dataset \cite{reddy2020analysis}. Especially when the training data set is small, the PCA-supervised discriminative approach can outperform; it is also less sensitive to the variability of the training sets \cite{martinez2001pca}.  The study \cite{garate2020classification} shows that PCA can increase the performance of different ML classifiers for the prediction of cardiac failure.

It can be said that the classifiers performed better after applying LDA to the linear data set. So, in the case of linear data, LDA can reduce the dimensionality and be used in different classification tasks \cite{ghosh2019improving}. However, the TF-IDF enhanced with the LDA approach did not allow the classifier to score high accuracy compared to the other two methods when smaller datasets were fed \cite{dzisevivc2019text}. One of the reasons was explained in \cite{reddy2020analysis}; the results showed that ML algorithms with PCA produce better results when the dimensionality of the data sets is high. When the dimensionality of datasets is low, the ML algorithms without dimensionality reduction yield better results. Another possible way is using an unsupervised generative Latent Dirichlet allocation to estimate the topic distribution (topics) by using observed variables (words). Latent Dirichlet allocation shows the effectiveness of overcoming the sparsity from the feature space matrix of TF-IDF \cite{kim2019research}. It can also help to make texts more semantically focused and reduce sparseness \cite{chen2016short}. However, its selection of characteristics does not improve performance with small data \cite{fodeh2019classification}. 

We explored the possibility of PCA for sparsity reduction because of their advantages mentioned above. The training was tuned and performed, and the best performance was achieved with a decrease to 2 principal dimensions. The completed test has an accuracy of 88\%; this is less than the performance of SelectKBest with 89\%. Furthermore, following the recommendation of \cite{laghmati2020classification}, we also tested a statistical method, Neighborhood Component Analysis (NCA) \cite{goldberger2005neighbourhood}, to reduce the dimensions of the data set. NCA has shown that it works well on a small dataset for the medical domain. However, the result is slightly better than PCA; NCA only achieves an accuracy of 89\% (the same as SelecKBest). Consequently, neither PCA nor NCA can improve the classification accuracy. It confirms the limitation of these approaches by linearly approximating a feature subspace to maximize class separability.

% From Fig. \ref{fig:PCA}, \ref{fig:NCA} we can easily see the overlapping of the features; hence the classification task hardly separates the boundary for the binary classification.

% \begin{figure}[!t]
% 	\centering
% 	\vspace{2pt}
% 	\includegraphics[scale=0.6]{photo/PCA.png}
% 	\vspace{-2pt}
% 	\caption{Visualization of the representation space for 2 components from Principle Component Analysis (PCA).}
% 	\vspace{-5pt}
% 	\label{fig:PCA}
% \end{figure}

% \begin{figure}[!t]
% 	\centering
% 	\vspace{2pt}
% 	\includegraphics[scale=0.6]{photo/NCA.png}
% 	\vspace{-2pt}
% 	\caption{Visualization of the representation space for 2 components from Neighborhood Component Analysis (NCA).}
% 	\vspace{-5pt}
% 	\label{fig:NCA}
% \end{figure}

Furthermore, non-linear activation function AE (AE) shows its best performance on compression the sparse TF-IDF representation space. We compare the effectiveness of reconstruction based on the reconstruction evaluation from Eq. \ref{eq:reconst_eval} between PCA, linear activation function AE (LAE), AE, and stacked AE (SAE) \cite{gehring2013extracting}. The results confirm that the PCA and LAE have the same performance, achieving about 80\% of the reconstruction. When the activation of AE is linear, then PCA and LAE are identical. In addition, there is no improvement if the SAE is used to extract the features in cases of limited data. Besides, the effectiveness of non-linear activation in AE is proved, when it can maximally reconstruct up to 86\% compared to the original spare data. It is one of the advantages of nonlinear transformation from AE, which is trained by a neural network, superior to linear transformation from conventional approaches.

\begin{figure}[t]
	\centering
	\vspace{2pt}
	\includegraphics[scale=0.4]{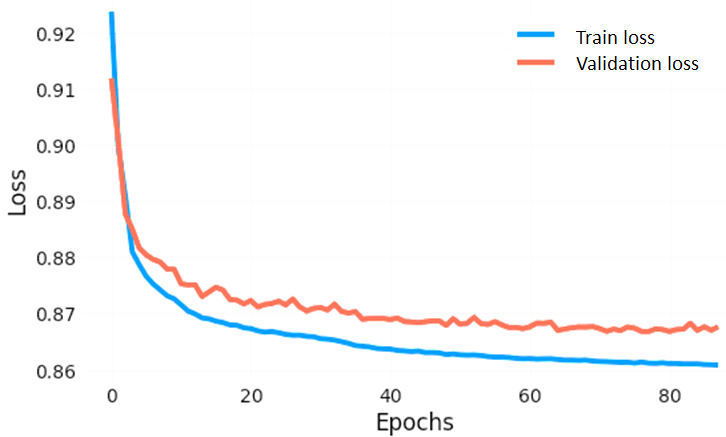}
	\vspace{-2pt}
	\caption{Loss for training and validation for the AE algorithm.}
	\vspace{-5pt}
	\label{fig:Learning_curves}
\end{figure}

\begin{figure}[!tp]
	\centering
	\includegraphics[scale=0.62]{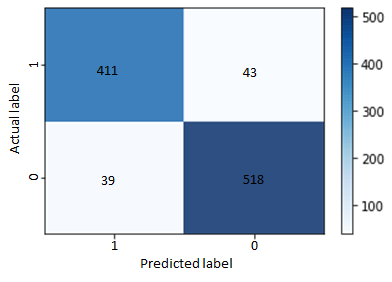}
	\caption{Confusion matrix of the MLP-NN classifier, showing the classification of positive (1) and negative (0) between predicted and actual labels.}
	\label{fig:confusion_matrix_J2}
\end{figure}

\begin{figure*}[!t]
	\centering
	\vspace{2pt}
	\includegraphics[scale=0.41]{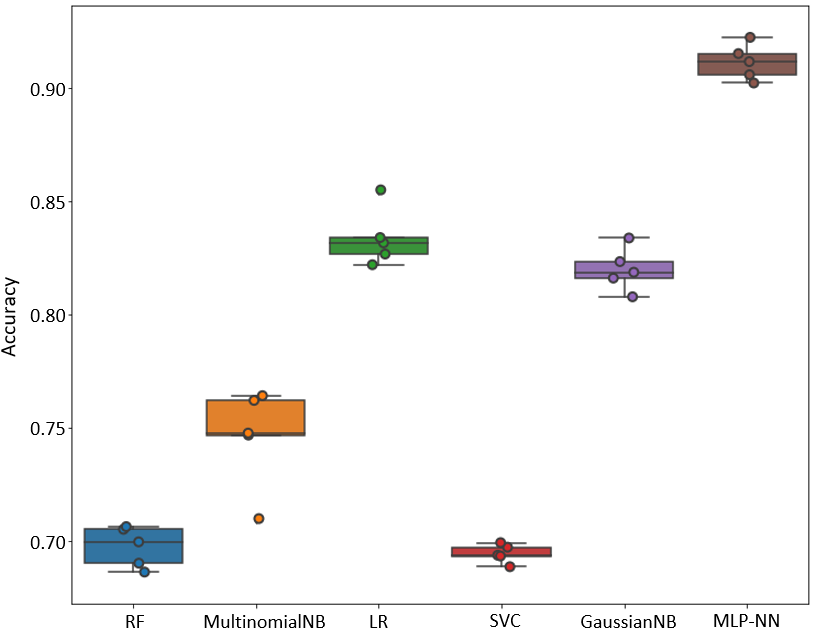}
	\vspace{-2pt}
	\caption{ A comparison evaluation of the box plot 5-fold cross-validation results for classifiers performance.}
	\vspace{-5pt}
	\label{fig:classifiers_comparison}
\end{figure*}

Overall, the downstream classification performances are effectively improved by feeding the compressed feature space output from the AE to ML classifiers. Fig. \ref{fig:Learning_curves} shows the loss during the training and validation process by optimizing the loss function from Eq. \ref{eq:reconstruction_error} for training the AE; the loss converges perfectly. After successfully training the AE, we have a pre-trained compressed, low-dimension feature space. We continue to the machine learning classifiers to perform the classification and evaluate the performance. Instead of performing on MLP-NN, LR, and GaussianNB, we also tested with other classifiers such as Random Forest (RF), Multinomial Naive Bayes, and Support Vector Machine. Fig. \ref{fig:classifiers_comparison} shows the comparison, using a box plot, of the 5-fold cross-validation. Again, MLP-NN gives the best performances; LR follows right after; GaussianNB is comparatively similar to LR. And all other classifiers are less effective. The best performance from MLP-NN is achieved at 92\%, 91\%, 91\% and 91\%, respectively, for accuracy, precision, recall, anf f1 score. And, the detailed confusion matrix showing the classification of positive cases (1) and negative cases (0) between predicted and actual labels is shown in Fig. \ref{fig:confusion_matrix_J2}. The experimental results are improved to 2-3 \% for each evaluation criterion from \cite{le2021detecting}, which had a general classification performance in a sparse TF-IDF feature space at 89\% accuracy, 88\% recall, and 89\% precision. It confirms that the AE method can deal with the sparsity by compressing the TF-IDF feature space. Consequently, it finally improves the downstream task performance of the MLP-NN classifier, and is also more robust than other methods. Recent work \cite{mienye2020improved} also confirmed a similar effect, but it was applied to a different dataset type and larger data availability. These results confirm the effectiveness of compressing the feature representation learning space into a low-dimensional representation using the AE algorithm. Notably, the robust transformation can outplay the deep learning models with limited data resources.

% \begin{figure}[t]
% 	\centering
% 	\vspace{2pt}
% 	\includegraphics[scale=0.675]{photo/Reconstruction_Evaluation.png}
% 	\vspace{-2pt}
% 	\caption{Data reconstruction evaluation comparison.}
% 	\vspace{-5pt}
% 	\label{fig:compress_compa}
% \end{figure}

% \begin{figure}[t]
% 	\centering
% 	\vspace{-5pt}
% 	\includegraphics[scale=0.8]{photo/GaussianNB_learning_curve.png}
% 	\vspace{-2pt}
% 	\caption{Performance of GaussianNB classifier in case of increasing the training size.}
% 	\vspace{-5pt}
% 	\label{fig:GNB_learning_curve}
% \end{figure}

% \begin{figure}[t]
% 	\centering
% 	\vspace{2pt}
% 	\includegraphics[scale=0.8]{photo/MLP_NN_learning_curve.png}
% 	\vspace{-2pt}
% 	\caption{Performance of MLP-NN classifier in case of increasing the training size.}
% 	\vspace{-5pt}
% 	\label{fig:MLPNN_learning_curve}
% \end{figure}

% \begin{figure}[!t]
% 	\centering
% 	\includegraphics[scale=0.8]{photo/ROC.png}
% 	\caption{ROC curve of MLP-NN classifier}
% 	\label{fig:ROC_curve}
% \end{figure}

\begin{figure*}[!t]
	\centering
	\includegraphics[scale=0.47]{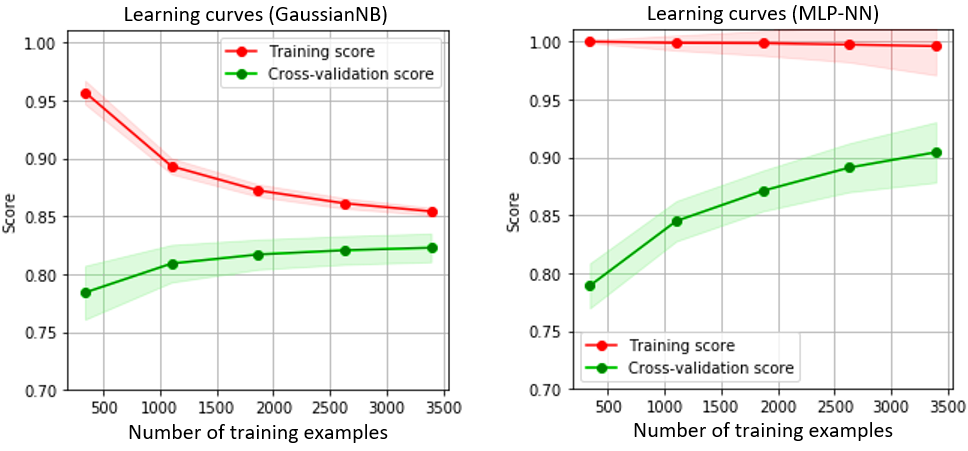}
	\caption{Performance of classifiers in case of increasing the training size: GaussianNB (left) and MLP-NN (right).}
	\label{fig:comparison_Gauss_MLP}
\end{figure*}

Furthermore, an important aspect of performance analysis is that the proposed approach still shows its advantageous capacity to increase data availability. We investigated the effectiveness of AE for compressing feature space, and studied how algorithm performance varies with the increasing of training examples from the compressed feature space. We assessed the performance of two classifiers GaussianNB and MLP-NN to evaluate the effectiveness. When it possibly increases data availability in the future, whether the classifier improves performance or not. In this case, study \cite{jordan2002discriminative} confirms that when the number of training examples increases, the generative model based on Naive Bayes would expect to perform better. However, our results are in contrast to that confirmation. Fig. \ref{fig:comparison_Gauss_MLP} shows the GaussianNB training and validation score when increasing the number of training examples. Technically, the GaussianNB reaches a plateau of performance after around 2000 training examples with the same dataset size, and the cross-validation score could not improve. We should expect that this is one of the limitations of GaussianNB, namely the linear discrimination characteristic for a real-world dataset, discussed in \cite{xue2008comment}. In contrast, the MLP-NN shows improvement with the increasing size of the dataset. Its cross-validation score gradually increases and shows no signs of reaching a maximum point. In short, we can confirm that when data is possibly increased, our approach still perfectly works and continually improves its performance for the downstream classification. %But, the capacity of the model should be adjusted to support the required fitting times.

% Yet, as shown in Fig. \ref{scalability_comp}, GaussianNB model is having a faster rate, where its fitting times are much smaller than the fitting time in the MLP-NN model. The fitting times of GaussianNB are nearly linear with the number of training examples. In comparison, the MLP-NN shows a trend of reaching a downward curve after a certain amount of training examples.

% \begin{figure*}[t]
% 	\centering
% 	\vspace{2pt}
% 	\includegraphics[scale=0.85]{photo/Scalability_comp.png}
% 	\vspace{-2pt}
% 	\caption{Scalability between GaussianNB and MLP-NN.}
% 	\vspace{-5pt}
% 	\label{scalability_comp}
% \end{figure*}

\begin{figure}
\includegraphics[scale=0.57]{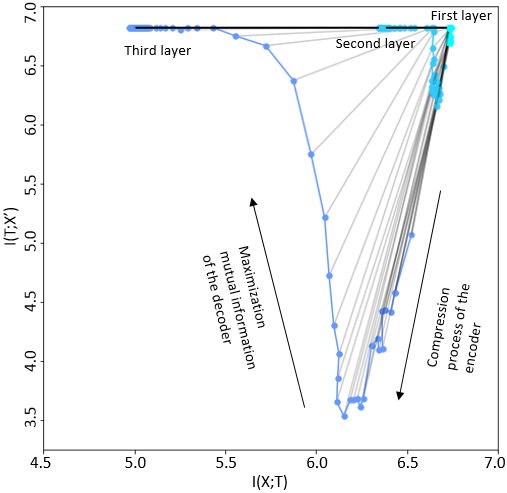}\\
\vspace{-2pt}
\includegraphics[scale=0.59]{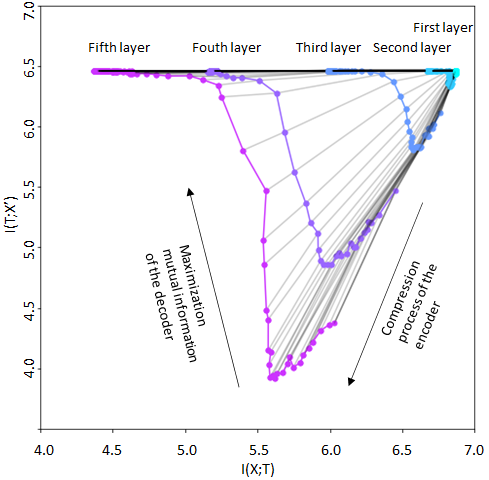}
\vspace{-2pt}
\caption{The evolution of the layers with epochs in the information plane for three hidden layers (top) and five hidden layers (bottom).}
\label{fig:MI_AE}
\end{figure}

Moreover, the behavior of AE in limited data is in harmony with more significant data cases based on the information-theoretic framework. We analyzed the behavior of AE based on an information-theoretic framework, as mentioned in Eq. \ref{eq:ib1}, and \ref{eq:ib2}. We want to understand how the AE behaves during the compression process by analyzing the mutual information of each hidden layer from the encoder and decoder. Generally, this type of analysis has been performed for a larger data set and has mainly focused on other data sources compared to our case; such as computer vision \cite{viola1997alignment}, medical imaging \cite{pluim2003mutual}, and genetics \cite{olsen2008impact}. We performed the analysis for two AE models with respecting to varying number of hidden layers (three hidden layers, and five hidden layers). As shown in Fig. \ref{fig:MI_AE}, there are two phases of the information plane in each hidden layer of the three-layer and five-layer cases. It is noted that from left to right it illustrates the behavior for each hidden layer. And in each hidden layer, from top to bottom, it captures the mutual information for each training epoch. Finally, all trajectories seem to follow a similar path during the learning process, then eventually converge and get closer to the optimal points in the theoretic information bottleneck bound.

Specifically, it can be divided into two phases for the working mechanism of AE in Fig. \ref{fig:MI_AE}. The first phase is called the drift phase, where the AE attempts to learn the latent representation $T(X)$ with a smaller dimension compared to the original data $X$. During the compression, there will be information lost, that is why we can see the trend of decreasing the mutual information of encoder $I(X; T)$. At the end of this step, we will have a compressed latent representation $T(X)$, and an optimal mutual information $I(X; T)$. Then, the second phase is named the diffusion phase. Within this step, the AE tries to find the reconstructed data $X'$, which are optimally close to the original data $X$. The AE maps the latent representation $T(X)$ to the reconstructed data $X'$ by maximizing the mutual information of the decoder $I(T; X')$. By doing that, we can see the increasing trend of $I(T; X')$; until $I(T; X')$ reaches its optimal bound for each layer.  One more interesting point is that the optimal mutual information will get smaller, when AE has more hidden layers. In case of three hidden layers, the optimal mutual information of the encoder $I(X, T)$ is larger by 6.0, but is maximum at 5.5 for five hidden layers. It is the same for the optimal mutual information of the decoder $I(T, X')$ at nearly 7.0 and 6.5 for three hidden layers and five hidden layers, respectively. These results illustrate the mechanism of an AE is to optimize the information bottleneck trade-off $T(X)$ during compression and prediction, respectively, for each layer. Remarkably, it is trained on a small and sparse dataset; still, it proves its effectiveness by compressing and maximizing the mutual information from the TF-IDF feature space.

% \begin{figure}[t]
% 	\centering
% 	\vspace{2pt}
% 	\includegraphics[scale=0.56]{photo/MI_AE_3.png}
% 	\vspace{-2pt}
% 	\caption{The evolution of the layers with the training epochs in the information plane with three hidden layers.}
% 	\vspace{2pt}
% 	\label{fig:MI_AE_3}
% \end{figure}
% \begin{figure}[t]
% 	\centering
% 	\vspace{2pt}
% 	\includegraphics[scale=0.55]{photo/MI_AE.png}
% 	\vspace{-2pt}
% 	\caption{The evolution of the layers with the training epochs in the information plane with five hidden layers.}
% 	\vspace{-5pt}
% 	\label{fig:MI_AE_5}
% \end{figure}

% \begin{figure*}[!htbp]
% 	\centering
% 	\includegraphics[scale=0.52]{photo/IB_Illustration.png}
% 	\caption{The evolution of the layers in the information plane with three hidden layers (left), and five hidden layers (right).}
% 	\label{fig:MI_AE}
% \end{figure*}

\section{Conclusion}
\label{sec:conclusion}

First, this study has shown that the participation of an AE in training can effectively compress the feature space of TF-IDF. The AE with a nonlinear activation function can achieve the reconstruction capacity at 86\% compared to the original data. It outperforms other approaches such as PCA, NCA, LAE (AE with linear activation function), and stacked AE. It concludes that AE can learn the best representation of the training data due to its lossless compression capacity.

Additionally, the AE also works well with a small clinical dataset; especially, in harmony with the information-theoretic mechanism of an AE for a larger dataset, and from different data sources. It has two phases of learning: the drift phase of the encoder by trying to compress the data. The second phase is related to the diffusion phase by maximizing the mutual information process in the decoder. Consequently, it shows the effectiveness of losses information in compressing the data. By doing so, we also capture the interpretability, comprehensibility, and transparency of the proposed model for decision making in our CDSS system as recommended by \cite{rudin2019stop}.

The second stage involves the use of an MLP-NN to predict the health status based on the compressed feature space. It has also been shown that the sparsity reduction for the feature space strongly affects the classifier performance in downstream task. AE learning algorithm effectively leverages the sparsity reduction. As a result, it helps the MLP-NN classifier achieve 92\% accuracy, 91\% recall, 91\% precision, and 91\% f1-score. This efficient ensemble model can outperform all alternative approaches: GaussianNB, LR, RF, MultimonialNB, and SVC.

The proposed approach is still proving successful in cases where data availability is increased. The MLP-NN is effective in achieve a better performance after the GaussianNB reaches its maximum capacity. In future work, we will choose the optimal parameters and validate our method on more datasets. We will explore the weak supervision approach that recently proved its effectiveness in 4,000 cardiac magnetic resonance sequences with imperfect labels \cite{fries2019weakly}; because it can maximize the use of unlabeled data at scale, which is costly to annotate.

\section*{Acknowledgment}

Clinical data were provided by the Research Center of CHU Sainte-Justine hospital, University of Montreal. The authors thank Dr. Sally Al Omar,  Dr. Michaël Sauthier, Dr. Rambaud Jérôme and Dr. Sans Guillaume for their data support of this research. This work was supported by a scholarship from the Fonds de recherche du Québec – Nature et technologies (FRQNT) to Thanh-Dung Le, and the grants from the Natural Sciences and Engineering Research Council (NSERC), the Institut de valorisation des données (IVADO), and the Fonds de la recherche en sante du Quebec (FRQS).

%\clearpage
\bibliographystyle{IEEEtran}
\bibliography{IEEEabrv,Bibliography}

\end{document}